\documentclass[10pt,journal,compsoc]{IEEEtran}
\usepackage{amsmath,amsfonts}
\usepackage{algorithmic}
\usepackage{algorithm}
\usepackage{array}
\usepackage[caption=false,font=normalsize,labelfont=sf,textfont=sf]{subfig}
\usepackage{textcomp}
\usepackage{stfloats}
\usepackage{bbding}
\usepackage{url}
\usepackage{verbatim}
\usepackage{graphicx}
\usepackage{float}

\usepackage{hyperref}
\usepackage{graphicx}

\usepackage[top=0.4in,bottom=0.4in,left=0.4in,right=0.4in]{geometry}
\usepackage{pifont}
\usepackage{cite}
\usepackage{color}
\usepackage{xcolor}
\usepackage{diagbox}
\usepackage{enumitem}
\usepackage{colortbl}  

\hypersetup{colorlinks=true,
    linkcolor=magenta,
    anchorcolor=blue,
    citecolor=blue,
    urlcolor=magenta
    }
\usepackage{makecell}
\usepackage{booktabs}
\usepackage{multirow}
\usepackage{orcidlink}
\usepackage{tikz}
\usepackage{array}
\usepackage{ulem}
\usepackage{ragged2e}
\usepackage{utfsym}
\usepackage{wasysym}
\usepackage{amssymb}  
\usepackage{threeparttable}
\usepackage{capt-of} 
\begin{document}


\title{HuiYanEarth-SAR: A Foundation Model for High-Fidelity and Low-Cost Global Remote Sensing Imagery Generation }


\author{Yongxiang Liu\hspace{-1.5mm}$^{~\orcidlink{0000-0002-0682-8365}}$, Jie Zhou
\hspace{-1.5mm}$^{~\orcidlink{0009-0004-3384-0556}}$, Yafei Song\hspace{-1.5mm}$^{~\orcidlink{0009-0004-6952-1071}}$, Tianpeng Liu
\hspace{-1.5mm}$^{~\orcidlink{0009-0004-3384-0556}}$, 
Li Liu
\hspace{-1.5mm}$^{~\orcidlink{0000-0002-2011-2873}}$



\IEEEcompsocitemizethanks{
			\IEEEcompsocthanksitem The authors are with the College of Electronic Science and Technology, National University of Defense Technology (NUDT), Changsha 410073, China. Email: lyx\_bible@sina.com, zhoujie\_@nudt.edu.cn, syf\_nudt@163.com, everliutianpeng@sina.cn, liuli\_nudt@nudt.edu.cn. This work was supported by National Natural Science Foundation of China (NSFC) under Grant 62376283 and 62531026, Innovation Research Foundation of National University of Defense Technology, and the Fundamental and Interdisciplinary Disciplines Breakthrough Plan of the Ministry of Education of China (JYB2025XDXM110).
             \IEEEcompsocthanksitem Corresponding authors: Yongxiang Liu and Li Liu
 		}
	}
 

\markboth{Journal of \LaTeX\ Class Files,~Vol.~14, No.~8, August~2021}%
{Shell \MakeLowercase{\textit{et al.}}: A Sample Article Using IEEEtran.cls for IEEE Journals}

\markboth{Submitted to IEEE TPAMI}{Jie Zhou}


\IEEEtitleabstractindextext{

\begin{center}\setcounter{figure}{0}
\centering
	\includegraphics[width=0.75\textwidth]{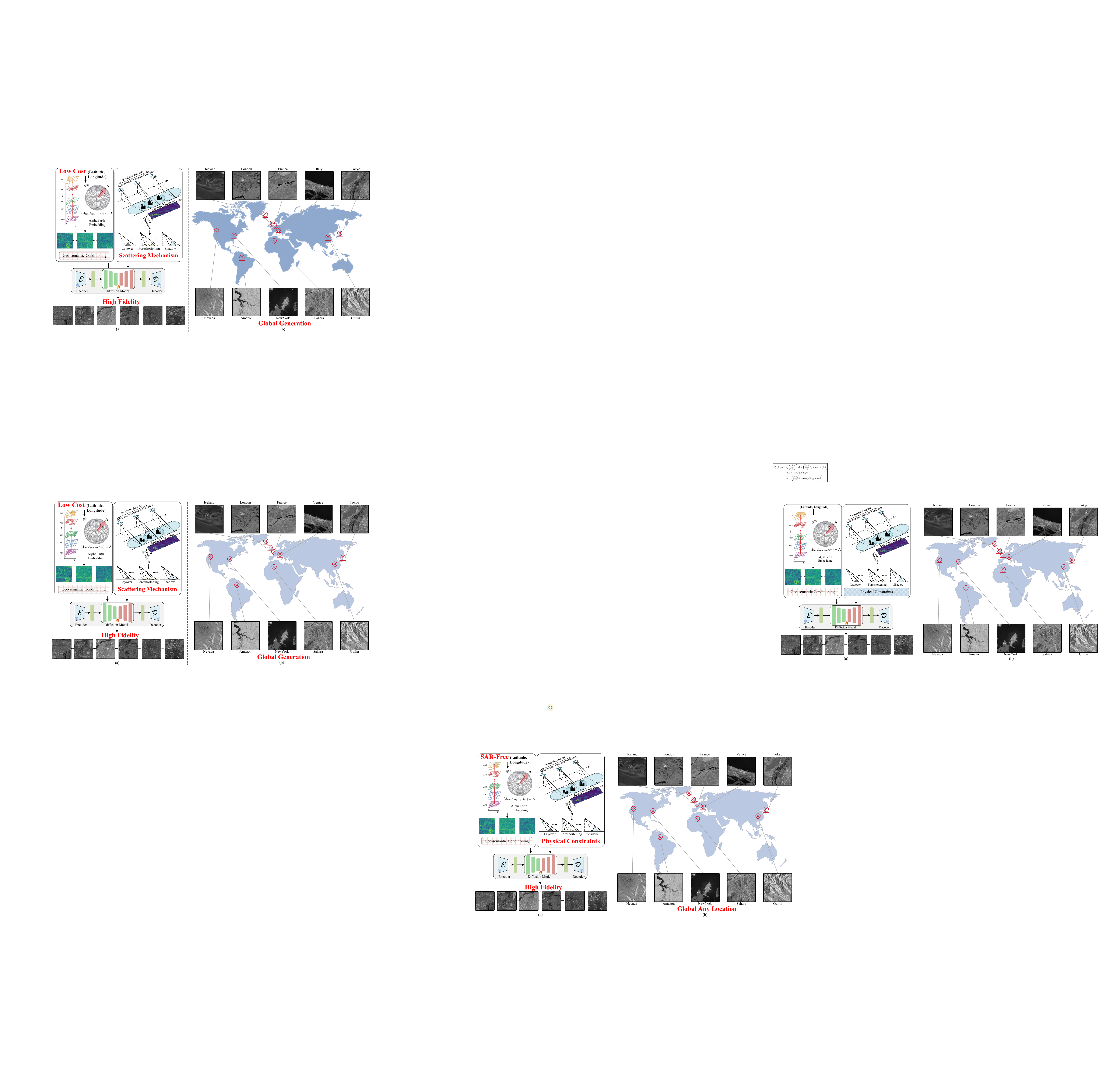}
	\captionof{figure}{(a) We propose HuiYanEarth-SAR, the first generative foundation model conditioned on both geophysical priors and SAR scattering mechanism. The model only takes the geographic coordinates as input with low cost, ultimately generating high-fidelity SAR images that are strictly aligned with the input conditions and physically consistent. (b) HuiYanEarth-SAR shows powerful capabilities in generating global, high-fidelity SAR images. Code is available at \href{https://github.com/JoyeZLearning/HuiYanEarth-SAR}{https://github.com/JoyeZLearning/HuiYanEarth-SAR}.}
 \label{abilityShow of HuiYanEarth-SAR}
 \end{center}

		\begin{abstract}
			\justifying
			{Synthetic Aperture Radar (SAR) imagery generation is essential for deepening the study of scattering mechanisms, establishing trustworthy electromagnetic scene models, and fundamentally alleviating the data scarcity bottleneck that constrains development in this field. However, existing methods find it difficult to simultaneously ensure high fidelity in both global geospatial semantics and microscopic scattering mechanisms, resulting in severe challenges for global generation. To address this, we propose \textbf{HuiYanEarth-SAR}, the first foundational SAR imagery generation model based on AlphaEarth and integrated scattering mechanisms. By injecting geospatial priors to control macroscopic structures and utilizing implicit scattering characteristic modeling to ensure the authenticity of microscopic textures, we achieve the capability of generating high-fidelity SAR images for global locations solely based on geographic coordinates. This study not only constructs an efficient SAR scene simulator but also establishes a bridge connecting geography, scatter mechanism, and artificial intelligence from a methodological standpoint. It advances SAR research by expanding the paradigm from perception and understanding to simulation and creation, providing key technical support for constructing a high-confidence digital twin of the Earth.  }
			
		\end{abstract}
		}

\maketitle
\IEEEdisplaynontitleabstractindextext
\IEEEpeerreviewmaketitle


\section{Introduction}
\IEEEPARstart{S}{ynthetic} Aperture Radar (SAR) possesses the unique advantage of all-day and all-weather Earth observation \cite{elachi1982spaceborne,zhou2025fifty,wang2026complex}, playing an irreplaceable role in fields such as environmental monitoring \cite{casagli2023landslide}, resource exploration \cite{dong2022chronic}, and disaster assessment \cite{sattar2025sikkim}. Advancing intelligent SAR interpretation technologies not only meets urgent practical demands for enhancing remote sensing information acquisition but also holds significant theoretical value in deepening the understanding of radar target scattering mechanisms and establishing trustworthy electromagnetic scene models. From a theoretical perspective, high-fidelity and large-scale SAR generative models can provide an unprecedented controlled data source and quantitative analysis tools for studying the electromagnetic scattering characteristics of various land covers, exploring the impact of imaging geometry and system parameters, and constructing radar signal simulation environments at the scene level \cite{liu2026atrnet}. From a practical standpoint, SAR data has always faced bottlenecks of limited scale and scarce annotations due to sensitivity and high acquisition costs. This has resulted in the current state where Artificial Intelligence (AI) technologies represented by Large Language Models and Multimodal Large Models cannot be fully empowered in the SAR domain as they are in optical remote sensing \cite{wang2025annotation,zhao2025remote,zheng2025Changen2}, with their potential shackled by data scarcity. Therefore, constructing a foundational model capable of generating high-fidelity SAR data is the key pathway to breaking this deadlock and achieving comprehensive AI empowerment for SAR interpretation.

However, SAR image generation faces unique and severe challenges, with an ecosystem far more complex than that of optical remote sensing. In the general remote sensing field, large scale image generation based on diffusion models has made great strides, with works emerging that can synthesize global optical imagery based on geographic coordinates \cite{khanna2023diffusionsat,yu2025Metaearth}. In contrast, the SAR domain lags behind, with existing research mostly focusing on the simulation and generation of specific targets \cite{zeng2024atgan,zhang2024shipGO,zhang2025iccvGAN}. While these works have conducted in-depth exploration at the target level, they are difficult to scale up to complex natural and urban scenes due to the complexity of scattering mechanism modeling and the scale of data-driven approaches, consequently failing to support the generation of large-scale SAR scenes at arbitrary global locations. The root cause lies in the complexity of SAR imaging mechanisms. The signal intensity and phase are coupled with multiple physical factors including sensor parameters, terrain geometry, and surface dielectric properties \cite{xiong2024qualitySARPAMI,zhang2025rsar,chen2024reinforcement}. This is further compounded by inherent speckle noise, creating a huge gap between visual plausibility and mechanistic correctness. Therefore, despite the urgent demand, works capable of reconciling macroscopic geospatial semantic consistency with microscopic scattering fidelity remain few and far between, constituting the core difficulty in this field.

Faced with the above challenges and opportunities, our research team, building upon a long-term commitment to SAR intelligent interpretation (e.g., our previously developed foundational model SAR ATR-X \cite{li2024saratrx} and the large-scale dataset ATRNet-STAR \cite{liu2026atrnet}), expands our research horizon from "target recognition" to "scene simulation." We recognize that constructing a SAR generative foundational model must break through the paradigm of "texture synthesis" and shift toward a new paradigm of "geography-physics" dual-driven generation. The recent emergence of geospatial foundation models AlphaEarth \cite{brown2025alphaearth} provides us with a key inspiration: their pre-trained geospatial embeddings can map any global coordinate into a semantic vector containing terrain, land cover, climate, and other priors, providing powerful conditional control for the generation process.

Based on this, we propose \textbf{HuiYanEarth-SAR}, a foundational model designed to generate high-fidelity SAR imagery at any global location under scattering mechanism constraints. Our core breakthrough lies in constructing a dual-driven generative framework guided by geospatial conditional priors and implicitly constrained by scattering mechanisms. Specifically, the model utilizes the AlphaEarth geospatial embedding as the core condition to ensure macroscopic semantic consistency between the generated scene and the input coordinates. Simultaneously, we integrate SAR scattering mechanisms (such as layover shadows and typical land cover scattering responses) and statistical characteristics (such as speckle noise) as implicit constraints into the diffusion model training through carefully designed noise schedules and loss functions. This ensures the microscopic fidelity of the generated results. Unlike traditional methods that require paired multi-modal data or complex physical simulations, our framework only needs geographic coordinates as input during the generation phase. It does not require any additional SAR data to output corresponding high-fidelity SAR images, providing a new pathway for global-scale and high-freedom SAR data synthesis and application.

The main contributions of this paper are summarized as follows:

(1) Propose a novel SAR generation paradigm based on geography-scattering mechanism synergy, systematically integrating AlphaEarth with SAR scattering constraints within a unified diffusion framework for the first time.

(2) Achieve high-fidelity global SAR imagery generation with low cost, only taking geographic coordinates as input, overcoming the limitations of existing methods in geographical coverage and fidelity.

(3) Design an implicit mechanism-aware training strategy that enables the diffusion model to autonomously learn and reproduce key SAR scattering characteristics without complex explicit physical simulations.

(4) Construct a holistic evaluation that comprehensively validates the high fidelity and practical value of the generated data through systematic qualitative analysis, quantitative metrics, hierarchical human vision assessments, and downstream task enhancement experiments.

\section{Related work}\label{relatedwork}

\subsection{General Visual Generative Models}

The evolution of image generation models has significantly advanced the field of computer vision. From early Generative Adversarial Networks (GANs) \cite{creswell2018GANReview} and Variational Autoencoders (VAEs) \cite{kingma2013VAE} to the recent dominance of diffusion models \cite{croitoru2023diffusion,zhou2025golden}, generative technologies have continuously progressed toward higher fidelity and stronger controllability. Among these, diffusion models have become the prevailing generative paradigm owing to their superior generation quality, training stability, and capability to model complex visual structures and semantic relationships \cite{chen2025invertibleDMPAMI}. DDPM \cite{ho2020DDPM} proposed a framework that learns forward noising and reverse denoising processes, laying the foundation for high-quality image synthesis. Subsequent improvements, such as DDIM \cite{song2020DDIM}, enhanced sampling efficiency, while LDM \cite{rombach2022SD} significantly reduced computational costs by performing denoising in a compressed latent space. Furthermore, DiT \cite{peebles2023DiT} replaced the U-Net architecture with Transformers, improving scalability and generation quality. These models have been widely applied to various conditional generation tasks, including text-to-image generation \cite{chen2024geodiffusion}, layout-to-image generation \cite{zhu2025objectsyn}, and others \cite{deltadahl2025deep,wang2026adaptive}, providing a solid technical foundation for controllable image generation in specialized domains such as remote sensing \cite{liu2024diffusionRSReview}.

\subsection{Remote Sensing Generative Foundation Models}

As the foundation model paradigm extends into the field of remote sensing, researchers have begun developing generative models capable of learning the universal distribution of Earth's appearance, supporting tasks such as data augmentation and scene simulation \cite{bodnar2025foundation,wu2025skysense}. Representative works such as MetaEarth \cite{yu2025Metaearth} have achieved global coordinate-based optical remote sensing image generation for arbitrary regions, demonstrating the potential for constructing a digital twin of the Earth. GeoSynth \cite{sastry2024geosynth} generates photorealistic satellite imagery by integrating OpenStreetMap data, while SatSynth \cite{toker2024satsynth} jointly produces images and semantic masks without external labels. HSIGene \cite{pang2025HSIGene} proposes a two-stage super-resolution framework to enable detail transfer from RGB to hyperspectral modalities. These methods have driven the transition of remote sensing generation from hand-crafted to data-driven. In the domain of change detection data synthesis, Changen2 \cite{zheng2025Changen2} alleviates the scarcity of multi-temporal annotated data through semantic label editing, whereas Noise2Change \cite{liu2025noise2change} generate temporally consistent yet semantically diverse image pairs, achieving a favorable balance between realism and alignment.

Despite these significant advances in optical remote sensing generation, SAR image generation faces unique challenges. The speckle noise, geometric distortions, and sensor-specific imaging mechanisms make it difficult to directly transfer most optical-image-based methods, leaving substantial room for exploration in this direction.

\subsection{SAR Image Generation Models}

The generation of SAR images is more challenging due to the unique coherent speckle noise and complex electromagnetic scattering mechanisms \cite{reigber2012very}. Early research primarily relied on Generative Adversarial Networks (GANs) \cite{zeng2024atgan,zhang2025iccvGAN}, alleviating the data scarcity issue in SAR image generation to some extent. However, their generation scope is typically limited to specific target categories and relies on restricted viewing angles. Recently, diffusion model-based SAR image generation has shown promise in data augmentation for object recognition and detection \cite{qosja2024sar,zhang2024shipGO}. For instance, Cross-Sensor SAR Generation achieves style transfer from existing sensors to new ones through LoRA fine-tuning and attention distillation, effectively mitigating data scarcity \cite{xuanting2025cross}. Nevertheless, existing SAR generation methods still face significant limitations. On one hand, most existing approaches are limited to unconditional or weakly conditional generation paradigms. These methods primarily focus on replicating the statistical textures of specific targets or scenes, rendering them inadequate for controllable generation at arbitrarily specified global locations. On the other hand, these techniques often process SAR data in isolation, failing to integrate with multi-source geospatial information such as optical imagery, elevation data, and climate records. Consequently, the generated images lack the geographical context consistency that is essential for real-world fidelity. Therefore, effectively combining SAR's imaging mechanisms with global-scale continuous geospatial semantics to achieve controllable and trustworthy image simulation remains a critical challenge to be addressed

\begin{figure}[!tb]
\includegraphics[width=0.45\textwidth]{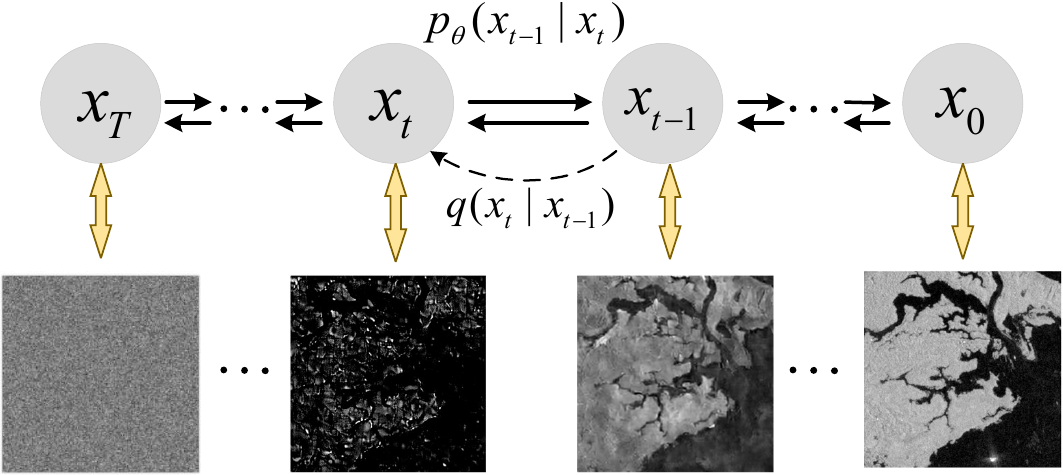}
\caption{Simplified schematic diagram of generation process.
}
\label{generation process}
\end{figure}

\section{Methodology}\label{methods}

This section elaborates on the architectural design and training methodology of the \textbf{HuiYanEarth-SAR}. Our primary objective is to establish a generative foundation model that takes arbitrary geographic coordinates (longitude and latitude) as input and generates SAR images with high geographic and  fidelity. The overall framework is illustrated in Fig. \ref{abilityShow of HuiYanEarth-SAR} (a). Its core innovation lies in a dual driving mechanism combining geographic condition guidance and scattering mechanism constraints. This design ensures that the model learns not only how to draw an SAR image but also what an SAR image should look like under specific geographical environments and scattering mechanism.

\subsection{Geo-semantic Condition}
To anchor the generation process within real geospatial space, we introduce AlphaEarth \cite{brown2025alphaearth} as a prior knowledge base and construct an end-to-end conditional encoding and injection pipeline.

\textit{(1) Geospatial Embedding Extraction and Condition Construction:} Given a target geographic coordinate, we utilize the pre-trained AlphaEarth Foundations model on Google Earth Engine (GEE) to extract its corresponding 64-dimensional global embedding vector $A=\left\{A_{00}, A_{01},...,A_{63} \right\}$. This vector has been proven to densely encode comprehensive information regarding the location's terrain, land cover, and climate zoning. During training and inference, we concatenate this embedding vector with the SAR imaging angle map (normalized to [-1, 1]) along the channel dimension to form a 65-dimensional conditional tensor $C^{65}$.

\textit{(2) Conditional Injection and Network Adaptation:} To effectively integrate conditional information into the generation process, we employ the ControlNet architecture for strong conditional control. Specifically, we expand the input channel count of the original layer from 3 to 65. The weights for the new channels are initialized by averaging and then replicating the original weights along the input channel dimension, thereby preserving the coherence of the network behavior. This ControlNet receives the conditional tensor $C^{65}$ and outputs multi-scale feature maps, which are injected into the encoder portion of the denoising U-Net in a residual form, achieving hierarchical guidance from macro to micro during the generation process.

\textit{(3) Efficient Fine-Tuning Strategy:} To efficiently adapt to SAR data while preserving the powerful generative priors of Stable Diffusion, we apply LoRA \cite{hu2021lora} fine-tuning to the U-Net backbone. LoRA modules are applied to the attention layers with the rank set to 32. During forward propagation, the low-rank increment produced by LoRA is added to the original weights: $\mathbf{W^{'}} = \mathbf{W} + \mathbf{BA}$, where $\mathbf{B}\in \mathbb{R}^{d\times r}$, $\mathbf{A}\in \mathbb{R}^{R\times k}$. This significantly reduces the number of trainable parameters, enabling data-efficient model customization.

\subsection{SAR Scattering Mechanism Constraints}

To enhance the physical realism of the generated images, we did not design additional physics-aware network modules. Instead, we improved the training objective to enable the model to implicitly yet effectively internalize the core scattering mechanisms and radiometric distribution patterns of SAR during end-to-end learning.

\textit{(1) Scattering Mechanism and Radiometric Distribution Modeling:} The essence of SAR imagery is the backscatter response of surface targets to radar waves. Its intensity distribution is closely related to geometric deformations such as layover, shadow, and foreshortening, as shown in Fig. \ref{abilityShow of HuiYanEarth-SAR}. For instance, mountain slopes facing the radar appear abnormally bright due to the layover effect, while slopes facing away appear extremely dark due to shadowing. Urban buildings generate strong double-bounce scattering due to the corner reflector effect. These phenomena collectively result in highly asymmetric and large dynamic range intensity distributions in SAR images. To prompt the generative model to master this complex radiometric distribution, we introduced offset noise during the denoising process of training. This technique forces the model to adjust the overall radiometric level while recovering local textures by superimposing a sample-level constant offset noise onto the standard Gaussian noise. This mechanism effectively enhances the model's representation capability for the coexistence of extremely bright areas (such as cities and radar-facing slopes) and extremely dark areas (such as water bodies and shadowed regions) in SAR imagery.

With initial state $x_{0}$, the forward process $q$ adds Gaussian noise $\epsilon$ over $T$ steps according to the noise schedule $\alpha_{t}$, where $t\in [1,2,...,T]$. The forward process is defined as
\begin{equation}
\begin{aligned}
x_t=\sqrt{\alpha_t} x_{t-1}+\sqrt{1-\alpha_t} \epsilon ; \epsilon \sim \mathcal{N}(\epsilon ; \mathbf{0}, \mathbf{I}),
\end{aligned}
\end{equation}

\begin{equation}
\begin{aligned}
q\left(x_t \mid x_{t-1}\right)=\mathcal{N}\left(x_t ; \sqrt{\alpha_t} x_{t-1},\left(1-\alpha_t\right) \mathbf{I}\right),
\end{aligned}
\end{equation}

\begin{equation}
\begin{aligned}
q\left(x_{1: T} \mid x_0\right)=\prod_{t=1}^T q\left(x_t \mid x_{t-1}\right),
\end{aligned}
\end{equation}
where the final state $x_{T}$ follows a Gaussian normal distribution with $x_{T}\sim \mathcal{N}(0,I)$.

The reverse process $p$ can be described as:
\begin{equation}
\begin{aligned}
p\left(x_{0: T}\right)=p\left(x_T\right) \prod_{t=1}^T p_\theta\left(x_{t-1} \mid x_t\right).
\end{aligned}
\end{equation}

Each step of the denoising process is learned by a neural network parameterized by $\theta$ and can be simplified as:
\begin{equation}
\begin{aligned}
p_\theta\left(x_{t-1} \mid x_t\right)=\mathcal{N}\left(x_{t-1} ; \mu_\theta\left(x_t, t\right), \Sigma_\theta\left(x_t, t\right)\right).
\end{aligned}
\end{equation}

Specifically, based on the standard Gaussian noise $\epsilon \sim \mathcal{N}(0,1)$ , we add a sample-wise shared offset noise:
\begin{equation}
\begin{aligned}
\epsilon^{'}=\epsilon+\gamma\cdot \nu, 
\end{aligned}
\end{equation}
where $\nu \sim \mathcal{N}(0,1)$ is constant across spatial dimensions, and $\gamma$=0.2 serves as the intensity coefficient. This forces the model to not only eliminate local detail noise during denoising but also learn to adjust the overall radiometric level of the image, thereby better modeling the extremely bright and dark regions common in SAR imagery, as shown in Fig. \ref{generation process}.

\textit{(2) Speckle Noise Statistical Constraints:} To strengthen the model's capability to generate SAR specific details, such as structural textures arising from complex scattering and multiplicative speckle noise, we adopted a Min-SNR weighting strategy in the loss function. This strategy dynamically assigns weights to the loss based on the signal-to-noise ratio at each step of the diffusion process, directing the model to focus more on learning high frequency details and complex structures that are difficult to reconstruct. This effectively guides the model to perform more precise modeling of geometric effect regions, such as layover and shadow, as well as the statistical characteristics of speckle noise during the generation process. The weight calculation is as follows:

\begin{equation}
\begin{aligned}
\text{SNR}(t)=\frac{\alpha_{t}}{1-\alpha_{t}},
\end{aligned}
\end{equation}

\begin{equation}
\begin{aligned}
\quad \text{weights}=\min \left(\text{SNR}(t), \gamma_{\mathrm{snr}}\right) / \text{SNR}(t),
\end{aligned}
\end{equation}
$\gamma_{\mathrm{snr}}$=5.0. The weighted loss function is formulated as follows:

\begin{equation}
\begin{aligned}
\mathcal{L}=\text { mean }\left(\text { weights } \cdot\left\|\epsilon-\epsilon_{\theta}\right\|_{2}^{2}\right),
\end{aligned}
\end{equation}
effectively strengthening the model's capability to generate typical SAR textures, such as granular speckle patterns.

Through the synergy of these two implicit scattering mechanism constraints, the model is able to automatically learn from the data and reproduce the key scattering phenomena and radiometric characteristics found in SAR imaging.



\section{Experiments}\label{experimentsetting}

\subsection{Experimental Setup}

\textit{(1) Dataset:} To achieve precise mapping between geographic locations and SAR signals, we constructed a globally multi-modal dataset with strict spatiotemporal alignment. Using the AlphaEarth geographic projection as the spatial reference, Sentinel-1 GRD data and Sentinel-2 Harmonized data were uniformly reprojected and resampled to 10-meter resolution based on geographic coordinates. For each geographic patch, we constructed a 71-dimensional multi-modal tensor with the following channel composition: (i) Geographic Embedding Source (4-D): This layer includes the Sentinel-2 B2 (Blue), B3 (Green), B4 (Red), and B8 (NIR) bands. It is primarily used for multi-modal alignment verification and downstream task expansion and does not participate directly in the gradient updates of the SAR generation model. (ii) SAR Image Source (3-D): This layer contains the VV polarization intensity, VH polarization intensity, and incidence angle. The data underwent radiometric calibration and terrain correction, with numerical ranges preserved as original decibel values to characterize the surface backscatter properties. (iii) Geospatial Embedding Layer (64-D): Composed of the pre-extracted 64-dimensional feature vector from AlphaEarth, this embedding layer densely encodes terrain, landform, climate zone, and seasonal vegetation index information for the location.

A sliding window strategy was adopted to crop the aligned large-scale imagery into 512$\times$512 patches, with a stride of 256 to achieve 50 percent overlapping sampling for data augmentation. During training, we implemented a dual-modal cleaning mechanism to remove invalid samples where the SAR channel was entirely black (edge data) or the embedding channel was missing, ensuring the physical consistency of the training data.

\textit{(2) Implementation Details:} We utilize the AdamW optimizer with an initial learning rate 2$\times$$10^{-5}$ and employ a cosine annealing schedule. The training is conducted on 8 NVIDIA 4090D GPUs, using mixed precision (fp16) to accelerate training and save memory. The parameters of the ControlNet and the LoRA modules in the U-Net are jointly optimized. During training, gradient checkpointing is also enabled to further reduce memory consumption. The final model supports the input of arbitrary latitude and longitude coordinates to generate corresponding 10-meter resolution Sentinel-1 SAR images. More details of the experimental setup can be found on our \href{https://github.com/JoyeZLearning/HuiYanEarth-SAR}{GitHub}.

\subsection{Qualitative Analysis}
 
\begin{figure*}[!tb]
\centering
\includegraphics[width=0.99\textwidth]{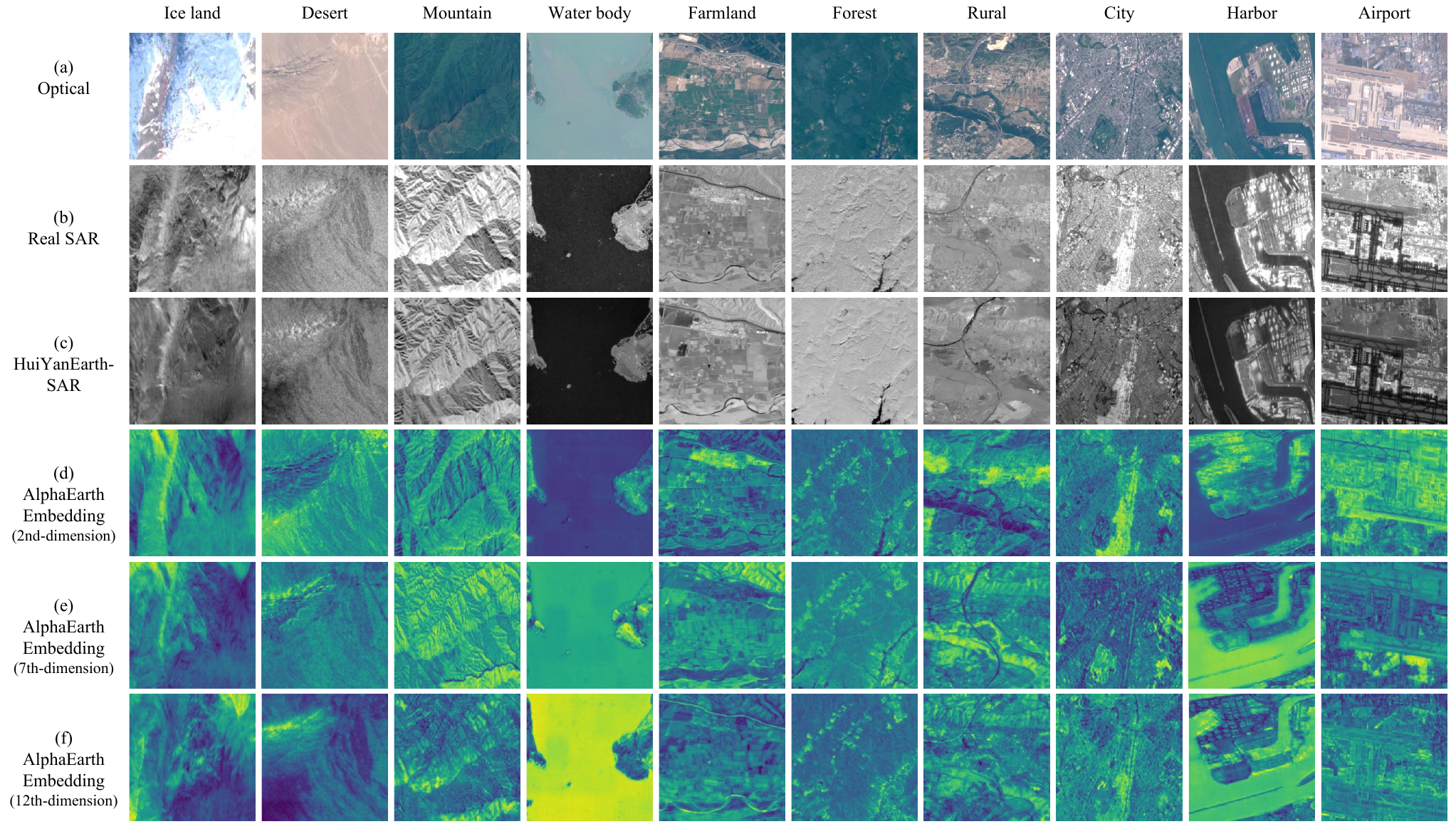}
\caption{Global diverse scene images generated by \textbf{HuiYanEarth-SAR}. (a) the optical reference image, (b) the real Sentinel-1 SAR image, and (c) the SAR image generated by our model under the same geographic coordinates. (d)-(f) show the visualization of the 2nd, 7th, and 12th dimensions of the AlphaEarth 64-dimensional embedding vector used in the input conditions, reflecting geographical semantic information at different levels. 
}
\label{multiscene generated by HuiYanEarth-SAR}
\end{figure*}

\textit{(1) Global-Scale Image Generation:} Fig. \ref{multiscene generated by HuiYanEarth-SAR} illustrates the generation performance of HuiYanEarth-SAR across ten globally representative scene categories. The generated SAR images exhibit a high degree of alignment with real SAR imagery in terms of macroscopic layout. For instance, the ridgelines and layover shadows in mountainous areas, as well as the piers and coastlines in harbor scenes, are accurately reproduced. This demonstrates that the AlphaEarth embedding effectively transfers geospatial semantic information. Regarding scattering mechanism characteristics, the generated images also display reasonable scattering responses: water bodies appear uniformly dark, urban areas exhibit bright speckled textures, vegetated regions show medium intensity with speckle noise, and airport runways manifest as dark linear strips. Furthermore, the images contain rich and natural details, such as furrows in farmland and internal shading variations within forests, without suffering from blurring or blocky artifacts. This quality is attributed to the LoRA fine-tuning strategy and noise optimization. Visualization of the AlphaEarth embedding further indicates that different dimensions correspond to distinct land cover features, suggesting that the model can decode abstract geospatial semantics into pixel-level representations that adhere to scattering mechanism.

\begin{figure}[!tb]
\centering
\includegraphics[width=0.495\textwidth]{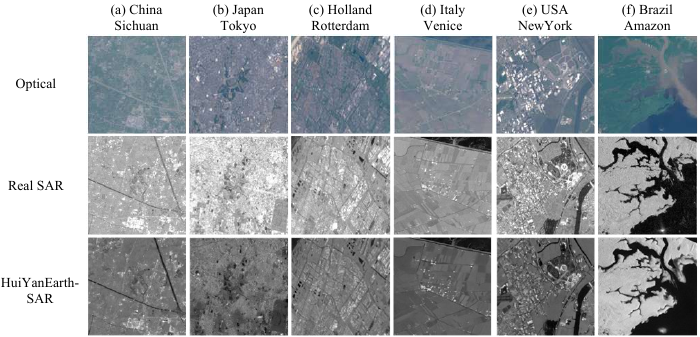}
\caption{Cross-continental multi-regional images generated by HuiYanEarth-SAR and their style consistency. Six representative locations are selected: (a) Sichuan China, (b) Tokyo Japan (Asia), (c) Rotterdam Netherlands, (d) Venice Italy (Europe), (e) New York USA, and (f) Amazon Brazil (Americas). For each column, from top to bottom, the images show the optical reference, the real SAR image, and the SAR image generated by our model. 
}
\label{multiregion generated by HuiYanEarth-SAR}
\end{figure}

Fig. \ref{multiregion generated by HuiYanEarth-SAR} further validates the generation capability of HuiYanEarth-SAR at a fine geospatial scale and its ability to capture regional styles across continents. The model adaptively generates SAR imagery reflecting typical characteristics of different continents, such as the water city pattern of Venice and the homogeneous texture of the Amazon rainforest. Meanwhile, the model distinguishes between urban and natural landscapes: urban scenes exhibit higher contrast and geometric regularity, whereas natural scenes present more uniform speckle noise. This indicates that the model has deeply learned the complex mapping relationship between diverse land cover types and SAR signal responses.

\begin{figure}[!tb]
\centering
\includegraphics[width=0.495\textwidth]{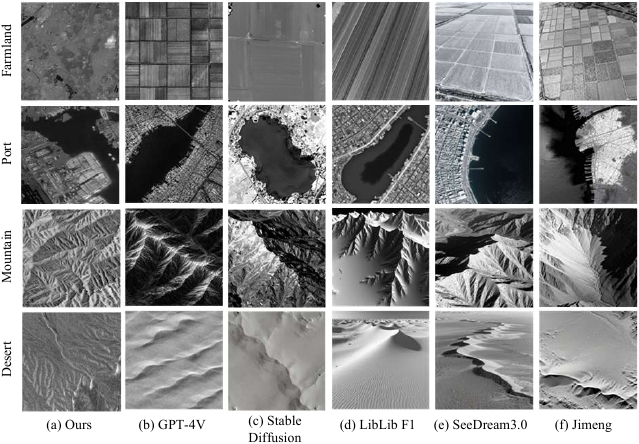}
\caption{Comparison between (a) HuiYanEarth-SAR and other text-to-image models, including (b) GPT-4V, (c) Stable Diffusion, (d) LibLib F1, (e) Seedream, and (f) Jimeng. The prompt words given to these models are all the same. Taking mountains as an example, the text prompts are givens as:"A 10-meter resolution Synthetic Aperture Radar (SAR) image of mountainous terrain. The image should show bright slopes facing the radar and dark slopes facing away due to layover and shadowing effects. The image has speckle noise and is in grayscale."
}
\label{Comparison with other T2I models}
\end{figure}

\textit{(2) Comparison With Other Text-to-Image Models:} Fig. \ref{Comparison with other T2I models} compares the performance of HuiYanEarth-SAR with mainstream text-to-image models on the SAR generation task. When prompted with the same professional descriptions (such as requiring mountain layover shadows and speckle noise), only our model accurately reflects the characteristics of SAR. It correctly renders geometric effects, such as bright slopes facing the radar and dark slopes facing away, along with physically plausible speckle noise textures. The generated results from other general-purpose models (GPT-4V \protect\footnote{\url{https://chatgpt.com/library}}, Stable Diffusion\protect\footnote{\url{https://stablediffusionweb.com/zh-cn}}, LibLib F1\protect\footnote{\url{https://www.liblib.art/}}, Seedream\protect\footnote{\url{https://seed.bytedance.com/zh/seedream3_0}}, and Jimeng\protect\footnote{\url{https://jimeng.jianying.com}}. ) exhibit clear limitations. These models fail to correctly interpret professional terminology. The generated images either lack SAR noise, present texture styles that resemble optical imagery and violate scattering mechanism, or even suffer from cartoonish distortions. This demonstrates that general-purpose models lack domain-specific knowledge and are unable to meet the requirements of professional generation tasks that demand strict geospatial and scattering mechanism constraints.

\subsection{Ablation Study}

\begin{figure}[!tb]
\centering
\includegraphics[width=0.495\textwidth]{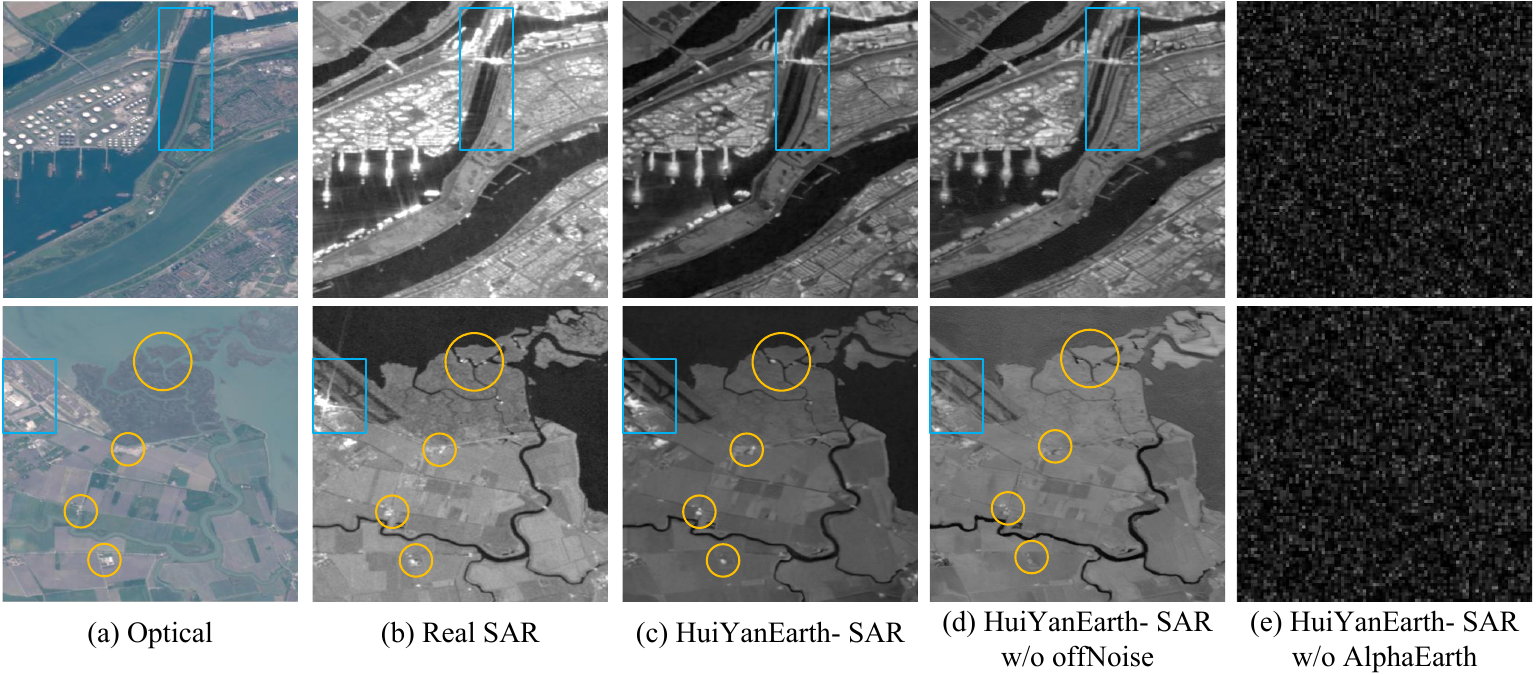}
\caption{Visual comparison of the core module ablation study. The blue bounding boxes highlight strong scatterers (e.g., bridges, large buildings). The complete model (c) accurately reconstructs these bright regions, whereas (d) appears significantly dimmer and blurred. The yellow circles indicate point targets, which are clearly visible in (c) but smoothed out or missing in (d). Column (e) demonstrates that the geospatial embedding is the fundamental prerequisite for generating valid imagery.
}
\label{Ablation study}
\end{figure}

In this section, we conduct experiments to validate the effectiveness of the proposed methods, as shown in Fig. \ref{Ablation study} and TABLE \ref{table:Ablation experiment}.

\textit{(1) The Necessity of Geographic Prior (AlphaEarth):} The most significant difference occurs in Fig. \ref{Ablation study} (e), Ours (w/o AlphaEarth Embedding). When the AlphaEarth embedding is removed and the model relies solely on coordinates or random noise as conditions, it fails completely, outputting meaningless pure noise. This result strongly proves that the diffusion model cannot directly map sparse spatial coordinates (latitude and longitude scalars) to complex SAR textures. The AlphaEarth embedding provides dense semantic information (e.g., water-land boundaries, urban layouts, vegetation distribution), serving as the critical condition to guide the diffusion process toward the correct geospatial structure. Without this prior, the model cannot distinguish land cover categories in the high-dimensional space.

\textit{(2) The Role of Scatter Mechanism Constraints:} Comparing Fig. \ref{Ablation study} (c), the complete model, with Fig. \ref{Ablation study} (d), the variant without scattering mechanism constraints, reveals the critical role of scattering mechanism modeling (specifically the noise shifting strategy and scattering mechanism loss) in enhancing image fidelity. As indicated by the blue bounding box covering bridges and large metal facilities, Fig. \ref{Ablation study} (c) successfully reproduces the strong secondary scattering effect typical of man-made metallic structures. In contrast, the variant Fig. \ref{Ablation study} (d) without the scattering constraint generates overly smoothed features in these regions, losing the distinct "sharpness" characteristic of SAR images. The yellow circles mark small point targets. In Fig. \ref{Ablation study} (c), these point targets maintain sharp and clear edges, whereas in Fig. \ref{Ablation study} (d), they are either drowned in or blurred by the background texture. By introducing the scattering constraint, our model is forced to preserve these high-intensity scattering peaks, thereby maintaining higher radiometric fidelity in such high-dynamic-range scenes.

\textit{(3) Quantitative Verification:} As shown in TABLE \ref{table:Ablation experiment}, the ablation study validates the core value of the dual-condition framework combining geographic and scattering mechanism constraints. When only the scattering mechanism constraint is enabled, the model completely fails, proving that geographic encoding is a necessary prerequisite for generating a reasonable scene structure. With only the geographic condition enabled, the model can generate images, but introducing the scattering constraint leads to significant improvements in radiometric distribution similarity, FSIM, and SSIM. This indicates that the scattering mechanism constraint significantly enhances micro-texture realism. The complete model achieves optimal performance across all metrics, corroborating the synergistic gain effect where geography guides the macro layout and the scattering mechanism constrains the micro textures, thus providing quantitative evidence for the dual-condition driven paradigm.

\begin{table}[!tb]
    \centering
        \caption{Ablation study results of core modules. Mean/Std Similarity measures radiometric distribution consistency between generated and real images. FSIM and SSIM assess structural similarity.}
\label{table:Ablation experiment}
    \begin{tabular}{cccccc}
    \toprule
        \multicolumn{2}{c}{Methods} &&  \multicolumn{3}{c}{Metrics} \\ \cline{1-2} \cline{4-6}
        AlphaEarth & Scattering && Mean/Std  & \multirow{2}*{FSIM$\uparrow$} & \multirow{2}*{SSIM$\uparrow$} \\ 
        Embedding& Mechanism && Similarity$\downarrow$ & ~ & ~ \\ \midrule
        $\times$ & $\times$ && - & - & -  \\ \hline
        $\times$& $\surd$  && - & - & - \\ \hline
        $\surd$  & $\times$ && 0.2028/0.0829 & 0.5693 & 0.3976 \\ \hline
       $\surd$  & $\surd$  &&\textbf{0.1752/0.0070} &  \textbf{0.6781}&\textbf{0.5625}\\ \bottomrule
    \end{tabular}
\end{table}

\subsection{Human Visual Evaluation}
To comprehensively evaluate the visual authenticity, geographical plausibility, and physical consistency of the generated images from a human perception perspective, we designed and conducted a hierarchical double-blind visual assessment experiment. This experiment aims to quantify the perceived realism of the generated images by experts at different cognitive levels and to provide subjective quality scores.

\textit{(1)  Experimental Design:} We selected 10 categories of representative global scenes (e.g., urban areas, forests, farmland, water bodies). Considering that skewed ratios (such as 1:3) are more suitable for evaluating human recognition capabilities in an environment inundated with generative content, we adopted a real: other generative models (in Sec. 4.2.2): our model ratio of 1:1:2, randomly selecting 12 images per session. The experiment employed an online double-blind questionnaire survey. Participants were pre-classified into three professional tiers, resulting in 24 valid questionnaires collected, as shown in TABLE \ref{table:Participant_grouping_information}. All participants were recruited through public academic channels and signed informed consent forms. 

\begin{table}[tb]
\begin{threeparttable} 
    \centering
    \caption{Participant information for the human visual evaluation.}
\label{table:Participant_grouping_information}
    \begin{tabular}{ccccc}
    \toprule
       \multirow{2}*{Level} & Personnel & Professional & Experience & Head- \\
        & Type & Background&  (Years)& count  \\   \midrule
        
        L1 & Novice  & with SAR basics & 1-3 & 18 \\ \hline
         \multirow{2}*{L2} &  \multirow{2}*{Familiar} & with SAR processing  &  \multirow{2}*{3–5} & \multirow{2}*{4} \\ 
         &   &   experience &   &  \\ \hline
        L3 & Expert & SAR image interpreters &$\ge$5 & 2 \\ \hline
        Total & - & - & - & 24 \\ \bottomrule
    \end{tabular}
    
      \begin{tablenotes} 
		\item * \textbf{Conflict of interest statement}: The authors of this study have no direct financial interest in AlphaEarth, Google, or other data providers. All evaluators were uninvolved in the development of this model or the annotation of the training data, ensuring the objectivity and impartiality of the assessment results. 
     \end{tablenotes} 
\end{threeparttable} 

\end{table}

\begin{figure}[!tb]
\centering
\includegraphics[width=0.495\textwidth]{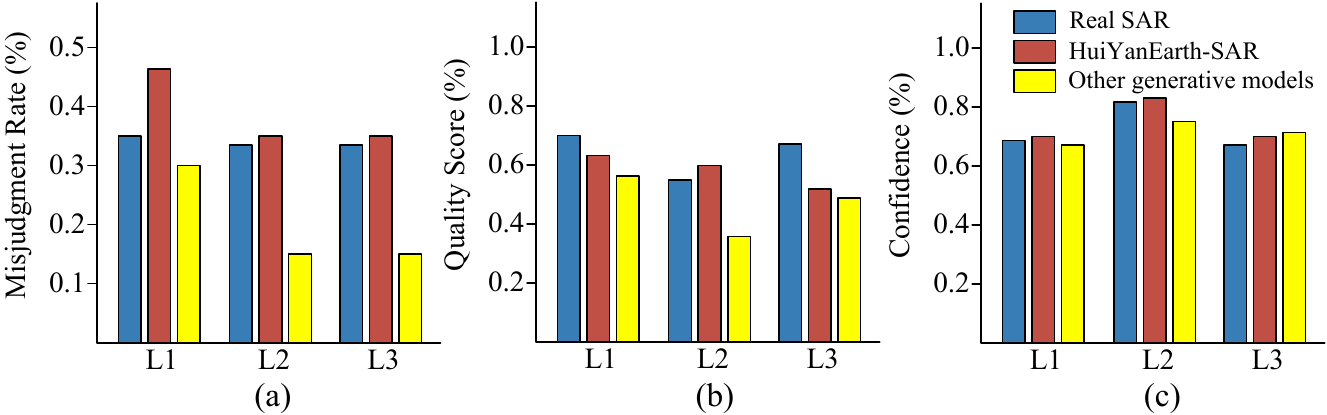}
\caption{Hierarchical human visual evaluation results. (a) Misjudgment rate. (b) Quality score. (c) Confidence. L1/L2/L3 denote novice, familiar, and expert groups, respectively. The results demonstrate that HuiYanEarth-SAR successfully achieves cross-level visual deception. The retention of approximately 35\% misjudgment even among experts proves that its generation quality approaches the limits of human visual discrimination.
}
\label{human vision Evaluation}
\end{figure}

\textit{(2) Experimental Procedure and Assessment Metrics:} Each participant was required to complete three tasks sequentially: (i) Authenticity Discrimination: Participants judged each image as either "real SAR" or "model-generated. " The misjudgment rate was calculated, specifically the proportion of generated images identified as real and the proportion of real images identified as generated within each group.
(ii) Quality Scoring: Regardless of their authenticity judgment, participants rated the overall realism, detail plausibility, and geographic consistency of the images. (iii) Confidence Self-Evaluation: Participants rated their confidence level regarding their own judgments.

\textit{(3) Result Analysis:} As shown in Fig. \ref{human vision Evaluation}, HuiYanEarth-SAR maintains a misjudgment rate of approximately 35\% among experts L3 which is comparable to real SAR 33\% and significantly higher than other generative models 15\%. The fact that even professional researchers struggle to distinguish the generated images proves that the model has transcended mere visual realism and achieved a new level of perceptual credibility. Moreover, the quality scores and confidence levels for the generated images are highly consistent with real SAR across all groups. Notably, in the L2 group, the quality score even surpasses that of real SAR. This indicates that the geographic-physical dual-constraint mechanism not only enhances objective fidelity but also delivers authentic subjective visual experiences.

\subsection{Evaluation on Downstream Tasks}
To quantitatively evaluate the practical value of HuiYanEarth-SAR as a data engine, we conducted downstream validation using scene classification, a fundamental remote sensing interpretation task. The experiment selected five typical scenes, including desert, forest, glacier, city, and mountain, with approximately 150 images of 512*512 size for each class. The dataset was split into training and testing sets at a ratio of 3:1. We employed four representative classifiers, VGG19, ResNet18, ResNet34, and ViT-B/32, to compare the performance differences between training on original data and with data augmentation.

As shown in TABLE \ref{table:downstream remote sensing image classification}, using HuiYanEarth-SAR-generated data to expand the training set fivefold resulted in consistent and significant performance improvements across all classifiers. This outcome validates the model's practical value from two aspects. First, the generated data is not merely a simple repetition but introduces effective feature variation, thereby increasing information richness and enabling the model to learn more robust class representations. Second, the geographic consistency and physical fidelity of the generated images successfully translate into performance gains for downstream tasks, demonstrating their potential to substitute real data.

From a broader perspective, the global generation capability of HuiYanEarth-SAR fosters unique applications. It can simulate multi-scale images with diverse regional characteristics, providing critical training resources for building remote sensing deepfake detection systems and contributing to the maintenance of geographic information security. In summary, this model not only alleviates the dilemma of data scarcity but also empowers downstream tasks in a low-cost and highly controllable manner, highlighting its core value as a SAR data engine.

\begin{table}[!tb]
    \centering
    \caption{Accuracy of the SAR scene classification task w/ and w/o augmentation.}
\label{table:downstream remote sensing image classification}
    \begin{tabular}{ccc}
    \toprule
        Methods & Training Data & Accuracy \\ \midrule
        \multirow{2}*{VGG16} & Real Data &91.26 \\ \cline{2-3}
        ~ & Augment Data & \textbf{95.37} \\ \hline
         \multirow{2}*{ResNet18} & Real Data & 92.35 \\\cline{2-3}
        ~ & Augment Data & \textbf{96.47} \\ \hline
        \multirow{2}*{ResNet34} & Real Data & 92.55 \\ \cline{2-3}
        ~ & Augment Data & \textbf{96.79} \\ \hline
         \multirow{2}*{ViT-B/32} & Real Data & 90.02 \\ \cline{2-3}
        ~ & Augment Data & \textbf{92.47} \\ \bottomrule
    \end{tabular}
\end{table}

\section{Limitations}

Despite the phased achievements of this work, as an exploratory study, its limitations outline a clear roadmap for future development. First, at the technical level, the current model is constrained by the underlying generative architecture, resulting in output resolution and detail richness that are insufficient for fine-grained characterization of ground objects, such as individual urban buildings. Concurrently, the model relies on static annual geographic embeddings and lacks explicit modeling capabilities for short-cycle, intense dynamic surface processes, such as flood evolution and rapid crop growth. Second, regarding the depth of physical modeling, the current scattering enhancement mechanism operates at the scene category level. The model's ability to generate and distinguish subtle scattering variations within the same land cover type caused by differences in material, structure, and moisture remains to be further strengthened.

\section{Conclusion}

In this paper, we introduced HuiYanEarth-SAR, a generative foundation model for global SAR imagery generation. By introducing AlphaEarth embeddings as geographic conditions and integrating SAR scattering mechanism modeling, the model achieves high-fidelity and low-cost SAR imagery generation for any location. Experiments demonstrate that this model surpasses existing methods in both generation fidelity and practicality, establishing itself as a credible SAR simulation and data augmentation platform. This research not only provides a pivotal technical solution to the scarcity of SAR data but also lays the foundation for building next-generation trustworthy remote sensing digital twin systems for Earth science research by tightly coupling data-driven approaches with physical laws.


\bibliographystyle{IEEEtran}
\normalem
\bibliography{cas-refs}

@article{li2024saratrx,
  title={{SARATR-X}: Toward Building A Foundation Model for {SAR} Target Recognition},
  author={Li, Weijie and Yang, Wei and Hou, Yuenan and Liu, Li and Liu, Yongxiang and Li, Xiang},
  journal={IEEE TIP}, 
  year={2025},
  volume={34},
  number={},
  pages={869-884},
  doi={10.1109/TIP.2025.3531988}
}

@article{xiong2024qualitySARPAMI,
  title={Quality improvement synthetic aperture radar ({SAR}) images using compressive sensing ({CS}) with Moore-Penrose inverse ({MPI}) and prior from spatial variant apodization ({SVA})},
  author={Xiong, Tao and Li, Yachao and Xing, Mengdao},
  journal={IEEE TPAMI},
  year={2024},
  publisher={IEEE}
}

@ARTICLE{yu2025Metaearth,
  author={Yu, Zhiping and Liu, Chenyang and Liu, Liqin and Shi, Zhenwei and Zou, Zhengxia},
  journal={IEEE TPAMI}, 
  title={{MetaEarth: A Generative Foundation Model for Global-Scale Remote Sensing Image Generation}}, 
  year={2025},
  volume={47},
  number={3},
  pages={1764-1781},
  doi={10.1109/TPAMI.2024.3507010}}

@ARTICLE{zheng2025Changen2,
  author={Zheng, Zhuo and Ermon, Stefano and Kim, Dongjun and Zhang, Liangpei and Zhong, Yanfei},
  journal={IEEE TPAMI}, 
  title={{Changen2: Multi-Temporal Remote Sensing Generative Change Foundation Model}}, 
  year={2025},
  volume={47},
  number={2},
  pages={725-741},
  doi={10.1109/TPAMI.2024.3475824}}

@article{brown2025alphaearth,
  title={{AlphaEarth Foundations: An embedding field model for accurate and efficient global mapping from sparse label data}},
  author={Brown, Christopher F and Kazmierski, Michal R and Pasquarella, Valerie J and Rucklidge, William J and Samsikova, Masha and Zhang, Chenhui and Shelhamer, Evan and Lahera, Estefania and Wiles, Olivia and Ilyushchenko, Simon and others},
  journal={arXiv preprint arXiv:2507.22291},
  year={2025}
}

@article{dong2022chronic,
  title={{Chronic oiling in global oceans}},
  author={Dong, Yanzhu and Liu, Yongxue and Hu, Chuanmin and MacDonald, Ian R and Lu, Yingcheng},
  journal={Science},
  volume={376},
  number={6599},
  pages={1300--1304},
  year={2022},
  publisher={American Association for the Advancement of Science}
}

@article{casagli2023landslide,
  title={{Landslide detection, monitoring and prediction with remote-sensing techniques}},
  author={Casagli, Nicola and Intrieri, Emanuele and Tofani, Veronica and Gigli, Giovanni and Raspini, Federico},
  journal={Nat. Rev. Earth Env.},
  volume={4},
  number={1},
  pages={51--64},
  year={2023},
  publisher={Nature Publishing Group UK London}
}

@article{sattar2025sikkim,
  title={{The Sikkim flood of October 2023: drivers, causes, and impacts of a multihazard cascade}},
  author={Sattar, Ashim and Cook, Kristen L and Rai, Shashi Kant and Berthier, Etienne and Allen, Simon and Rinzin, Sonam and de Vries, Maximillian Van Wyk and Haeberli, Wilfried and Kushwaha, Pradeep and Shugar, Dan H and others},
  journal={Science},
  volume={387},
  number={6740},
  pages={eads2659},
  year={2025},
  publisher={American Association for the Advancement of Science}
}

@article{elachi1982spaceborne,
  title={{Spaceborne synthetic-aperture imaging radars: Applications, techniques, and technology}},
  author={Elachi, Charles and Bicknell, Tom and Jordan, Rolando L and Wu, Chialin},
  journal={Proceedings of the IEEE},
  volume={70},
  number={10},
  pages={1174--1209},
  year={1982},
  publisher={IEEE}
}

@article{reigber2012very,
  title={{Very-high-resolution airborne synthetic aperture radar imaging: SP and applications}},
  author={Reigber, Andreas and Scheiber, Rolf and Jager, Marc and Prats-Iraola, Pau and Hajnsek, Irena and Jagdhuber, Thomas and Papathanassiou, Konstantinos P and Nannini, Matteo and Aguilera, Esteban and Baumgartner, Stefan and others},
  journal={Proceedings of the IEEE},
  volume={101},
  number={3},
  pages={759--783},
  year={2012},
  publisher={IEEE}
}

@article{wu2025skysense,
  title={{A semantic-enhanced multi-modal remote sensing foundation model for Earth observation}},
  author={Wu, Kang and Zhang, Yingying and Ru, Lixiang and Dang, Bo and Lao, Jiangwei and Yu, Lei and Luo, Junwei and Zhu, Zifan and Sun, Yue and Zhang, Jiahao and others},
  journal={Nat. Mach. Intell.},
  pages={1--15},
  year={2025},
  publisher={Nature Publishing Group UK London}
}

@article{bodnar2025foundation,
  title={{A foundation model for the Earth system}},
  author={Bodnar, Cristian and Bruinsma, Wessel P and Lucic, Ana and Stanley, Megan and Allen, Anna and Brandstetter, Johannes and Garvan, Patrick and Riechert, Maik and Weyn, Jonathan A and Dong, Haiyu and others},
  journal={Nature},
  pages={1--8},
  year={2025},
  publisher={Nature Publishing Group UK London}
}

@ARTICLE{liu2026atrnet,
	title={{{ATRNet-STAR}: A Large Dataset and Benchmark Towards Remote Sensing Object Recognition in the Wild}},
    journal={IEEE TPAMI}, 
	author={Yongxiang Liu and Weijie Li and Li Liu and Jie Zhou and Bowen Peng and Yafei Song and Xuying Xiong and Wei Yang and Tianpeng Liu and Zhen Liu and Xiang Li},
    year={2026},
    volume={},
    number={},
    pages={},
}

@article{croitoru2023diffusion,
  title={{Diffusion models in vision: A survey}},
  author={Croitoru, Florinel-Alin and Hondru, Vlad and Ionescu, Radu Tudor and Shah, Mubarak},
  journal={IEEE TPAMI},
  year={2023},
  publisher={IEEE}
}

@inproceedings{qosja2024sar,
  title={{SAR Image Synthesis with Diffusion Models}},
  author={Qosja, Denisa and Wagner, Simon and O'Hagan, Daniel},
  booktitle={2024 IEEE Radar Conference (RadarConf24)},
  pages={1--6},
  year={2024},
  organization={IEEE}
}

@inproceedings{rombach2022SD,
  title={{High-resolution image synthesis with latent diffusion models}},
  author={Rombach, Robin and Blattmann, Andreas and Lorenz, Dominik and Esser, Patrick and Ommer, Bj{\"o}rn},
  booktitle={CVPR},
  pages={10684--10695},
  year={2022}
}

@inproceedings{khanna2023diffusionsat,
  title={{Diffusionsat: A generative foundation model for satellite imagery}},
  author={Khanna, Samar and Liu, Patrick and Zhou, Linqi and Meng, Chenlin and Rombach, Robin and Burke, Marshall and Lobell, David and Ermon, Stefano},
  journal={ICLR},
  year={2024}
}

@article{wang2025annotation,
  title={{Annotation-Free, High-Fidelity SAR Oil Spill Image Synthesis via Classification-Guided Diffusion Model}},
  author={Wang, Bin and Dai, Song and Song, Dongmei and Chen, Lei and Chen, Weimin and Yu, Jintao},
  journal={IEEE TGRS},
  year={2025},
  publisher={IEEE}
}

@article{zhao2025remote,
  title={{Remote Sensing Image Generation via Object Text Decoupling}},
  author={Zhao, Wenda and Zhang, Zhepu and Zhao, Fan and Wang, Haipeng and He, You and Lu, Huchuan},
  journal={IEEE TPAMI},
  year={2025},
  publisher={IEEE}
}

@article{zhang2024shipGO,
  title={{Ship-Go: SAR ship images inpainting via instance-to-image generative diffusion models}},
  author={Zhang, Xin and Li, Yang and Li, Feng and Jiang, Hangzhi and Wang, Yanhua and Zhang, Liang and Zheng, Le and Ding, Zegang},
  journal={ISPRS JPRS},
  volume={207},
  pages={203--217},
  year={2024},
  publisher={Elsevier}
}

@article{liu2025noise2change,
  title={{Generating Any Changes in the Noise Domain}},
  author={Liu, Qiang and Kuang, Yang and Yue, Jun and Ghamisi, Pedram and Xie, Weiying and Fang, Leyuan},
  journal={IEEE TPAMI},
  year={2025},
  publisher={IEEE}
}

@article{creswell2018GANReview,
  title={{Generative adversarial networks: An overview}},
  author={Creswell, Antonia and White, Tom and Dumoulin, Vincent and Arulkumaran, Kai and Sengupta, Biswa and Bharath, Anil A},
  journal={IEEE SPM},
  volume={35},
  number={1},
  pages={53--65},
  year={2018},
  publisher={IEEE}
}

@article{kingma2013VAE,
  title={{Auto-encoding variational bayes}},
  author={Kingma, Diederik P and Welling, Max},
  journal={arXiv preprint arXiv:1312.6114},
  year={2013}
}

@article{liu2024diffusionRSReview,
  title={{Diffusion models meet remote sensing: Principles, methods, and perspectives}},
  author={Liu, Yidan and Yue, Jun and Xia, Shaobo and Ghamisi, Pedram and Xie, Weiying and Fang, Leyuan},
  journal={IEEE TGRS},
  year={2024},
  publisher={IEEE}
}

@article{ho2020DDPM,
  title={{Denoising diffusion probabilistic models}},
  author={Ho, Jonathan and Jain, Ajay and Abbeel, Pieter},
  journal={Advances in NeuIPS},
  volume={33},
  pages={6840--6851},
  year={2020}
}

@article{song2020DDIM,
  title={{Denoising diffusion implicit models}},
  author={Song, Jiaming and Meng, Chenlin and Ermon, Stefano},
  journal={arXiv preprint arXiv:2010.02502},
  year={2020}
}

@article{chen2025invertibleDMPAMI,
  title={{Invertible diffusion models for compressed sensing}},
  author={Chen, Bin and Zhang, Zhenyu and Li, Weiqi and Zhao, Chen and Yu, Jiwen and Zhao, Shijie and Chen, Jie and Zhang, Jian},
  journal={IEEE TPAMI},
  year={2025},
  publisher={IEEE}
}

@inproceedings{peebles2023DiT,
  title={{Scalable diffusion models with transformers}},
  author={Peebles, William and Xie, Saining},
  booktitle={ICCV},
  pages={4195--4205},
  year={2023}
}

@article{zhu2025objectsyn,
  title={{Object detection data synthesis via box-to-image generation based on diffusion models}},
  author={Zhu, Jingyuan and Ma, Huimin and Chen, Jiansheng and Yuan, Jian},
  journal={IEEE TPAMI},
  year={2025},
  publisher={IEEE}
}

@inproceedings{chen2024geodiffusion,
  title={{GeoDiffusion: Text-Prompted Geometric Control for Object Detection Data Generation}},
  author={Chen, Kai and Xie, Enze and Chen, Zhe and Wang, Yibo and Hong, Lanqing and Li, Zhenguo and Yeung, Dit-Yan},
  booktitle={ICLR},
  year={2024}
}

@inproceedings{zhou2025golden,
  title={{Golden noise for diffusion models: A learning framework}},
  author={Zhou, Zikai and Shao, Shitong and Bai, Lichen and Zhang, Shufei and Xu, Zhiqiang and Han, Bo and Xie, Zeke},
  booktitle={ICCV},
  pages={17688--17697},
  year={2025}
}

@inproceedings{wang2026adaptive,
  title={{Adaptive Domain Shift in Diffusion Models for Cross-Modality Image Translation}},
  author={Wang, Zihao and Chen, Yuzhou and Ren, Shaogang},
  booktitle={ICLR},
  year={2026}
}

@article{deltadahl2025deep,
  title={{Deep generative classification of blood cell morphology}},
  author={Deltadahl, Simon and Gilbey, Julian and Van Laer, Christine and Boeckx, Nancy and Leers, Mathie PG and Freeman, Tanya and Aiken, Laura and Farren, Timothy and Smith, Matthew and Zeina, Mohamad and others},
  journal={Nat. Mach. Intell.},
  volume={7},
  number={11},
  pages={1791--1803},
  year={2025},
  publisher={Nature Publishing Group UK London}
}

@inproceedings{sastry2024geosynth,
  title={{Geosynth: Contextually-aware high-resolution satellite image synthesis}},
  author={Sastry, Srikumar and Khanal, Subash and Dhakal, Aayush and Jacobs, Nathan},
  booktitle={CVPR},
  pages={460--470},
  year={2024}
}

@inproceedings{toker2024satsynth,
  title={{Satsynth: Augmenting image-mask pairs through diffusion models for aerial semantic segmentation}},
  author={Toker, Aysim and Eisenberger, Marvin and Cremers, Daniel and Leal-Taix{\'e}, Laura},
  booktitle={CVPR},
  pages={27695--27705},
  year={2024}
}

@ARTICLE{pang2025HSIGene,
  author={Pang, Li and Cao, Xiangyong and Tang, Datao and Xu, Shuang and Bai, Xueru and Zhou, Feng and Meng, Deyu},
  journal={IEEE TPAMI}, 
  title={{HSIGene: A Foundation Model for Hyperspectral Image Generation}}, 
  year={2026},
  volume={48},
  number={1},
  pages={730-746},
  doi={10.1109/TPAMI.2025.3610927}}

@article{xuanting2025cross,
  title={{Cross-Sensor SAR Data Generation Using Diffusion Models and Feature Migration}},
  author={Xuanting, WU and Fan, ZHANG and Fei, MA and Qiang, YIN and Yongsheng, ZHOU},
  journal={Trans. of Nanjing Uni. of Aero. \& Astro.},
  volume={42},
  number={4},
  year={2025}
}

@inproceedings{zhang2025iccvGAN,
  title={{$\mathbf{\Phi}$-GAN: Physics-Inspired GAN for Generating SAR Images Under Limited Data}},
  author={Zhang, Xidan and Zhuang, Yihan and Guo, Qian and Yang, Haodong and Qian, Xuelin and Cheng, Gong and Han, Junwei and Huang, Zhongling},
  booktitle={ICCV},
  year={2025}
}

@article{zeng2024atgan,
  title={{ATGAN: A SAR target image generation method for automatic target recognition}},
  author={Zeng, Zhiqiang and Tan, Xiaoheng and Zhang, Xin and Huang, Yan and Wan, Jun and Chen, Zhanye},
  journal={IEEE JSTARS},
  volume={17},
  pages={6290--6307},
  year={2024},
  publisher={IEEE}
}

@article{zhou2025fifty,
  title={{Fifty Years of SAR Automatic Target Recognition: The Road Forward}},
  author={Zhou, Jie and Liu, Yongxiang and Liu, Li and Li, Weijie and Peng, Bowen and Song, Yafei and Kuang, Gangyao and Li, Xiang},
  journal={arXiv preprint arXiv:2509.22159},
  year={2025}
}

@inproceedings{hu2021lora,
      title={LoRA: Low-Rank Adaptation of Large Language Models}, 
      author={Edward J. Hu and Yelong Shen and Phillip Wallis and Zeyuan Allen-Zhu and Yuanzhi Li and Shean Wang and Lu Wang and Weizhu Chen},
      booktitle={ICLR},
      year={2022}
}

@inproceedings{zhang2025rsar,
  title={{RSAR: Restricted state angle resolver and rotated SAR benchmark}},
  author={Zhang, Xin and Yang, Xue and Li, Yuxuan and Yang, Jian and Cheng, Ming-Ming and Li, Xiang},
  booktitle={CVPR},
  pages={7416--7426},
  year={2025}
}

@article{chen2024reinforcement,
  title={A reinforcement learning framework for scattering feature extraction and SAR image interpretation},
  author={Chen, Jiacheng and Zhang, Xu and Wang, Haipeng and Xu, Feng},
  journal={IEEE TGRS},
  volume={62},
  pages={1--14},
  year={2024},
  publisher={IEEE}
}

@article{wang2026complex,
  title={A Complex-valued SAR Foundation Model Based on Physically Inspired Representation Learning},
  author={Wang, Mengyu and Bi, Hanbo and Feng, Yingchao and Xin, Linlin and Gong, Shuo and Wang, Tianqi and Yan, Zhiyuan and Wang, Peijin and Diao, Wenhui and Sun, Xian},
  journal={IEEE TIP},
  year={2026},
  publisher={IEEE}
}






\vfill

\end{document}